\documentclass[oneside,12pt]{article}
\usepackage{geometry}
 \usepackage[left]{lineno}
\usepackage{booktabs,cite}
\usepackage[colorlinks]{hyperref}

\usepackage{graphicx,epstopdf}
\usepackage{amsopn}
\usepackage{bold-extra}
\usepackage{mathtools}
\usepackage{psfrag}
\usepackage{amssymb}
\usepackage{amsmath}
\usepackage{graphicx}
\usepackage{xcolor}
\usepackage{amssymb}
\usepackage{units}

\usepackage{xspace}
\usepackage{bold-extra}

\usepackage{wasysym}
\usepackage[color=green!40]{todonotes}

\colorlet{lred}{red!40}
\colorlet{lblue}{blue!40}
\colorlet{lgreen}{green!40}

\definecolor{mixc}{cmyk}{0.5,0.5,0.5,0}
\colorlet{mixl}{mixc!30}

\usepackage{enumitem}

\usepackage{enumitem}
\setitemize{label=\scriptsize{$\blacksquare$}}
\setenumerate{label=(\alph*)}
\usepackage{amssymb,amsthm}

\setenumerate{label=(\alph*)}

\numberwithin{equation}{section}
\numberwithin{theorem}{section}

\newcommand{\source}{h}
\newcommand{\hilbert}[1]{\mathcal{H}}

\newcommand{\R}{\mathbb R}

\newcommand{\edot}{\,\cdot\,}

\newcommand{\waved}{\mathcal{P}}
\newcommand{\wave}{\mathbf P}

\newcommand{\ph}{\varphi}

\newcommand{\rr}{r}
\newcommand{\pp}{\mathbf p}

\newcommand{\rmd}{\mathrm d}

\newcommand{\coloneqq}{:=}

\newcommand\sset[1]{\{#1\}}
\newcommand\abs[1]{\left\vert#1\right\vert}
\newcommand\sabs[1]{\lvert#1\rvert}
\newcommand\norm[1]{\left\Vert#1\right\Vert}
\newcommand\snorm[1]{\Vert#1\Vert}

\newcommand{\kl}[1]{\left(#1\right)}

\newcommand{\WW}{\mathcal W}
\newcommand{\TT}{\mathcal{T}}

   \DeclarePairedDelimiterX\set[1]\{\}{%
      
      #1
   }

\newcommand{\Xin}{\mathbf X}
\newcommand{\Xout}{\mathbf Y}
\newcommand{\Nx}{d}
\newcommand{\Nl}{L}
\newcommand{\Nf}{F}

\usepackage{soul}
\usepackage{xcolor}

\colorlet{lred}{red!40}
\colorlet{lgreen}{green!40}
\colorlet{lblue}{blue!40}
\definecolor{mixc}{cmyk}{0.5,0.5,0.5,0}
\colorlet{mixl}{mixc!30}

\title{Deep Learning for Photoacoustic Tomography from Sparse Data}

\author{Stephan~Antholzer   \and Markus~Haltmeier  \and Johannes~Schwab}

\date{Department of Mathematics, University of Innsbruck\\[0.2em]
Technikerstrasse 13, A-6020 Innsbruck, Austria\\[0.2em]
Corresponding e-mail: {\tt markus.haltmeier@uibk.ac.at}}

\begin{document}

\maketitle

\begin{abstract}
The development of fast and accurate image reconstruction algorithms is a central aspect of computed tomography.
In this paper, we investigate this issue for the sparse data problem in photoacoustic tomography (PAT). We develop a direct and highly efficient reconstruction algorithm based on deep learning.
In our approach image reconstruction is performed with a deep convolutional neural network (CNN), whose weights are adjusted prior to the actual image reconstruction based on a set of training data. The proposed reconstruction approach can be interpreted as a network that uses the PAT filtered backprojection algorithm for the first layer, followed by the U-net architecture for the remaining layers. Actual image reconstruction with deep learning consists in one evaluation of the trained CNN, which  does not require  time consuming solution of  the  forward and adjoint problems. At the same time, our numerical results demonstrate that the proposed deep learning approach reconstructs images with a quality comparable to state of the art iterative approaches for PAT from sparse data.

\noindent\textbf{keywords}
Photoacoustic tomography, sparse data,  image reconstruction, deep learning, convolutional neural networks,  inverse problems.

\end{abstract}

\section{Introduction}
\label{sec:intro}
Deep learning is a rapidly emerging research area  that  has significantly improved performance
of many pattern recognition and machine learning applications~\cite{lecun2015deep,goodfellow2016deep}.
Deep learning   algorithms make use of special artificial neural network designs
for representing a  nonlinear input to output map together with  optimization procedures
for adjusting the weights of the network during the training phase. Deep learning techniques are currently the  state of the art for visual object  recognition,  natural language understanding or applications in other domains such as
drug discovery or biomedical image analysis (see, for example, \cite{collobert2011natural,bahdanau2014neural,ma2015deep,krizhevsky2012imagenet,ioffe2015batch,wu2015deep,litjens2017survey,greenspan2016guest} and the references therein).

Despite its  success in various scientific disciplines, in  image reconstruction deep
learning research appeared only very recently (see \cite{chen2017lowdose,wang2016accelerating,han2016deep,jin2016deep,wang2016perspective,wurfl2016deep,zhang2016image}).
In this paper,  we develop a deep learning framework for image reconstruction in photoacoustic tomography (PAT).
To concentrate on the main ideas we restrict ourselves  to the sparse data problem in PAT in a circular measurement geometry.
Our approach can be extended to an arbitrary  measurement geometry in arbitrary dimension.
Clearly, the increased dimensionality   comes with an increased computational cost. This is especially the case for
 the training of the network which, however, is done prior to the actual image reconstruction.

\subsection{PAT and the sparse sampling problem}

\begin{figure}[tbh!]
\begin{center}
    \includegraphics[width=\columnwidth]{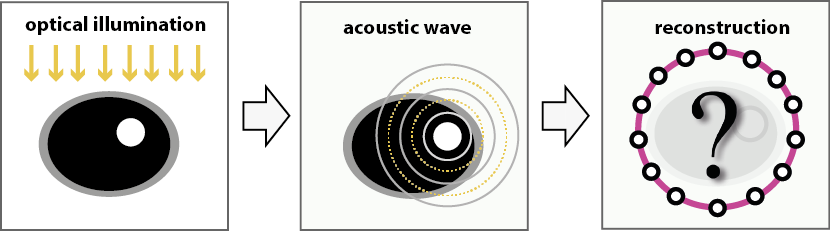}
\end{center}
\caption{\textsc{Basic\label{fig:pat} principle of PAT}. An object  is illuminated with a  short optical pulse (left) that induces an acoustic pressure wave (middle). The pressure signals are recorded outside of the object, and are used to
recover an image of the interior (right).}
\end{figure}

PAT  is a non-invasive  coupled-physics biomedical imaging technology  which beneficially combines the high  contrast of
pure optical imaging with  the high spatial resolution of ultrasound imaging \cite{beard2011biomedical,kruger1995photoacoustic,paltauf2007photacoustic,wang2009multiscale}. It is based on the generation of
acoustic waves by illuminating a  semi-transparent biological or medical object with short optical pulses.
The induced time dependent acoustic waves are measured outside of the sample with acoustic detectors,
and the measured data are used to recover an image of the interior (see Figure~\ref{fig:pat}).
High spatial resolution in PAT can be achieved
by measuring the acoustic data with high spatial and temporal  sampling rate \cite{haltmeier2016sampling,natterer2001mathematics}.
While temporal samples can be easily collected at or above the
Nyquist rate, the spatial sampling density is usually
limited~\cite{arridge2016accelerated,beuermarschallinger2015photacoustic,gratt201564line,rosenthal2013acoustic,guo2010compressed,sandbichler2015novel}. In fact, each spatial measurement requires a separate sensor and high quality
detectors are often costly.  Moving the  detector elements  can increase the  spatial sampling rate, but is time consuming and also can introduce motion artifacts.  Therefore, in actual  applications, the number of sensor locations is usually small compared to the desired resolution which  yields to a so-called sparse data problem.

Applying standard algorithms to sparse data yields low-quality images containing severe undersampling artifacts.  To some extent, these artifacts can be reduced by using iterative image reconstruction algorithms  \cite{arridge2016adjoint,belhachmi2016direct,haltmeier2017iterative,huang2013full,javaherian2017multi,schwab2016galerkin,wang2012investigation}  which allow to include prior knowledge such as smoothness, sparsity or total variation (TV) constraints \cite{daubechies2014sparsity,frikel2017efficient,grasmair2010sparsity,provost2009application,acar1994analysis,scherzer2009variational}.  These algorithms tend to be time consuming as the forward and adjoint problems have to be solved  repeatedly. Further, iterative algorithms have their own limitations. For example, the reconstruction quality strongly depends on the used a-priori  model about  the objects to be recovered.
For example,  TV minimization assumes sparsity of the gradient of the image to be reconstructed.
Such assumptions are often not strictly satisfied in real world scenarios which again limits the theoretically achievable reconstruction quality.

To overcome the above limitations, in  this paper we develop a new reconstruction approach based on deep learning that comes with the following properties:
(i)  image reconstruction is efficient and non-iterative;
(ii)  no explicit a-priori model for the class of objects to be reconstructed is required;
(iii) it yields a reconstruction quality comparable  to (or even outperforming) existing methods for sparse data.
Note that instead of providing an explicit a-priori model, the deep learning  approach requires a set of training data
and the CNN itself adjusts to the provided training data. By training the network on real word data, it thereby automatically creates a model of the considered PAT images in an implicit and purely data driven manner.  While training of the  CNN again requires  time consuming iterative minimization, we stress that training is  performed 
independent of the particular investigated objects  and  prior  to the  actual image reconstruction.  Additionally, if  the time resources  
available for  training  a new network are limited,  then one can  use 
weights learned on one  data set as good starting  value for training the weights 
in the new network.

\subsection{Proposed deep learning approach}
\label{sec:approach}

\begin{figure}[tbh!]
\begin{center}
    \includegraphics[width=\columnwidth]{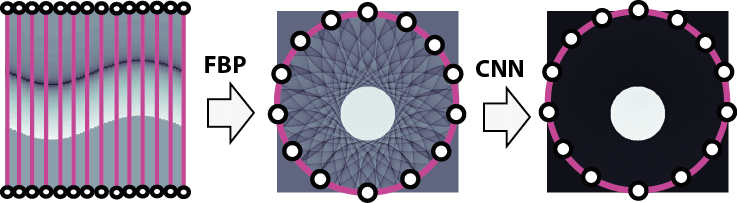}
\end{center}
\caption{\textsc{Illustration of the proposed network for PAT image reconstruction}.\label{fig:approach}
In the first step, the FBP algorithm (or another standard linear reconstruction method)  is applied to the sparse data. In a second step,  a deep convolutional neural network is applied to the intermediate reconstruction which outputs an almost  artifact free image. This may be interpreted  as a deep network with the FBP in the first layer and the CNN in  the remaining layers.}
\end{figure}

Our reconstruction approach for the sparse data problem in PAT  uses
a deep convolutional neural network in combination with
any linear reconstruction method as preprocessing step. Essentially, it consists of the
following two steps (see Figure~\ref{fig:approach}):
\begin{enumerate}[label=(D\arabic*)]
\item \label{step1} In the first step, a  linear PAT  image reconstruction algorithm
is applied, which  yields an approximation of the original object  including under-sampled artifacts.

\item \label{step2} In the second step,  a deep convolutional neural network (CNN)
is applied  to map the intermediate reconstruction from \ref{step1} to an artifact free final image.
\end{enumerate}
Note that  the above two-stage procedure can be viewed as a single deep neuronal network that uses  the linear reconstruction algorithm in \ref{step1} for the first layer, and the  CNN  in  \ref{step2} for the  remaining layers.

Step  \ref{step1} can be implemented  by any standard linear reconstruction
algorithm including  filtered backprojection (FBP)  \cite{Kun07a,finch2004determining,FinHalRak07,Hal13a,burgholzer2007temporal,xu2005universal,haltmeier14universal}, Fourier methods \cite{AgrKuc07,Kun07b,jaeger2007fourier,HalSchZan09b,XuXuWan02}, or
time reversal \cite{burgholzer2007exact,Treeby10,HriKucNgu08,SteUhl09}.
In fact, all these methods can be implemented efficiently using at most $\mathcal{O}(\Nx^3)$
floating point operations (FLOPS) for reconstructing  a high-resolution image on an $\Nx \times \Nx$ grid. 
Here $\Nx$ is number  spatial  discretization points  along  one dimension of the reconstructed image.   
The CNN  applied in  step \ref{step2} depends on weights that are adjusted using
a set of training data to  achieve artifact removal. The weights in the CNN are adjusted
during the so-called training  phase which is performed prior to the actual image reconstruction \cite{goodfellow2016deep}.
In  our current implementation, we use the U-net architecture  originally designed in \cite{ronneberger2015unet}
for image  segmentation.  Application of the trained network for image reconstruction is  fast. One application of the U-net requires  $\mathcal{O}(\Nf^2\Nl \Nx^2)$ FLOPS, where $\Nf$ is the number of channels  in the first convolution and $\Nl$ describes the depth of the network. Typically, $\Nf ^2\Nl$
will be in the order of $\Nx$ and the number of FLOPS for evaluating  the CNN is comparable  to the effort of performing
an FBP reconstruction.
Moreover, evaluation of the CNN can easily be parallelized, which further increases numerical performance.
On the other hand, iterative reconstruction algorithms tend to be slower as they require repeated application of the
PAT forward operator and its adjoint.

To the best of our knowledge, this is the first paper using deep learning or neural networks
for  PAT. Related approaches applying CNNs for different medical imaging
technologies including computed tomography (CT) and  magnetic resonance imaging (MRI)
appeared recently in \cite{wang2016perspective,wang2016accelerating,chen2017lowdose,jin2016deep,han2016deep,wurfl2016deep,zhang2016image}.
The author of \cite{wang2016perspective} shares his opinions on deep learning for image reconstruction. In \cite{jin2016deep}, deep learning is
applied to imaging problems where the
normal operator is shift invariant; PAT does not belong to this class.
A different  learning approach for addressing the limited view problem in PAT is proposed in~\cite{dreier2017operator}.
The above references  show that  a significant amount of research has been done on deep learning for  CT and MRI image reconstruction (based on inverse Radon and inverse Fourier transforms).    Opposed  to that, PAT requires inverting the   wave equation, and our work is the first paper that used  deep learning  and CNNs for PAT reconstruction and  inversion of the wave equation.

\subsection{Outline}

The rest of this paper is organized as follows. In Section~\ref{sec:pat} we  review PAT
and discuss the sparse sampling problem.  In Section~\ref{sec:deep} we describe the proposed  deep  learning approach. For that purpose, we discuss neural networks and present CNNs and the U-net  actually implemented in  our approach. Details on  the numerical implementation and
numerical results are  presented in Section~\ref{sec:num}. The paper concludes with a short
summary and outlook given in Section~\ref{sec:discussion}.

\section{Photoacoustic tomography}
\label{sec:pat}

As illustrated in Figure~\ref{fig:pat}, PAT is based
on generating an acoustic wave inside some investigated object using short optical pulses.
Let us denote by $\source \colon \R^d \to \R$ the initial pressure
distribution which provides diagnostic information about the
patient and which is the quantity of interest  in  PAT. Of practical relevance  are the
cases $d=2, 3$ (see \cite{kuchment2011mathematics,paltauf2007photacoustic,wang2006photoacoustic}).
For keeping the presentation simple and focusing on  the main ideas we only consider the  case of  $d=2$. 
 Among others, the two-dimensional case arises in PAT with so called integrating line detectors \cite{paltauf2007photacoustic,burgholzer2007temporal}.
 Extensions to three spatial dimensions are  possible by using the  FBP algorithm for 3D PAT \cite{finch2004determining}
in combination with the  3D U-net designed in \cite{cicek20163dunet}). 
Further, we restrict ourselves to the case of a circular measurement geometry, where the acoustic measurements are made on a circle  surrounding the investigated object.  In general  geometry, one can use the  so-called  universal backprojection 
formula \cite{xu2005universal,haltmeier14universal}  that is  exact for general  geometry up to an additive smoothing term \cite{haltmeier14universal}. In this case, the  CNN  can be used to account for the  under-sampling  issue as well as to 
account the  additive smooth term.
Such investigations, however, are beyond the scope of this paper.

\subsection{PAT in circular measurement geometry}

In two spatial dimensions, the induced  pressure in PAT satisfies the following initial value problem for the  2D wave equation
\begin{equation} \label{eq:wave}
\left\{\begin{array}{ll}
\partial^2_t p (x,t)  - \Delta p(x,t) = 0 & \text{ for } (x,t) \in \R^2 \times \kl{0,\infty} \\
p(x,0) = \source (x)   & \text{ for } x \in \R^2 \\
\partial_t p(x,0) = 0 & \text{ for } x \in \R^2 \,,
\end{array}\right.
\end{equation}
where we assume a constant sound-speed that is rescaled to one.
In the circular measurement geometry, the initial pressure $\source$ is assumed to vanish outside  the  disc
$B_R \coloneqq  \sset{x \in \R^2 \mid  \norm{x} < R}$. 
Note that the solution of used  forward wave equation \eqref{eq:wave} is, for positive times,  equal to the causal  solution of the wave equation with source term $\delta'(t) h(x)$; see
\cite{haltmeier2017new}.  Both models (either with source term or with initial condition) are frequently used in PAT. The goal of PAT image reconstruction is
to recover $\source$ from measurements of the acoustic pressure $p$ made on the boundary $\partial B_R$.

In a complete data situation, PAT in a circular measurement geometry consist in recovering the function $\source$ from data
\begin{equation} \label{eq:boundary-data}
    (\wave \source)  (z,t) \coloneqq  p(z,t) \quad \text{ for } (z,t) \in \partial  B_R  \times  [0,T] \,,
\end{equation}
where $p$ denotes the solution of~\eqref{eq:wave}
with initial data $\source$ and $T$  is the final measurement time.
Here complete data refers to data prior to sampling that are known  on the full  boundary $\partial  B_R $ and up to times  $T \geq 2R$. In such a case, exact  and  stable PAT image reconstruction is theoretically possible; see\cite{haltmeier2017iterative,stefanov2009thermoacoustic}.   
Several efficient methods for recovering $\source$ from complete data $\wave \source$ are well investigated (see, for example,  \cite{Kun07a,finch2004determining,FinHalRak07,Hal13a,burgholzer2007temporal,xu2005universal,haltmeier14universal,AgrKuc07,Kun07b,jaeger2007fourier,HalSchZan09b,XuXuWan02,burgholzer2007exact,Treeby10,HriKucNgu08,SteUhl09}). As an example, we mention the  FBP
formula  derived in \cite{FinHalRak07}, \begin{equation} \label{eq:fbp2d}
          \source(\rr)
         =
        - \frac{1}{\pi R}
        \int_{\partial B_R}
        \int_{\abs{\rr-z}}^\infty
        \frac{ (\partial_t t \wave \source)(z, t)}{ \sqrt{t^2-\sabs{\rr-z}^2}}  \, \rmd t
        \rmd S(z)
         \,.
\end{equation}
Note that \eqref{eq:fbp2d} requires data for all $t>0$; see~\cite[Theorem 1.4]{FinHalRak07}
for a related  FBP formula  that  only uses data for $t < 2R$.  For the numerical results in this paper we truncate \eqref{eq:fbp2d}
at $t=2R$, in which situation all singularities of the initial pressure are contained in the reconstructed
image and the truncation error is small.

\begin{psfrags}
\psfrag{p}{\scriptsize $\ph_k$}
\psfrag{m}{\scriptsize $r_\ell$}
\psfrag{f}{}
\begin{figure}[htb!]
\begin{centering}
\includegraphics[width=0.8\columnwidth]{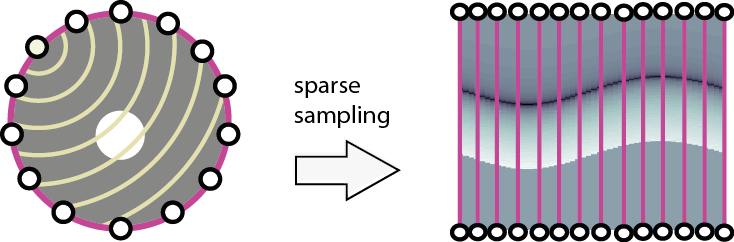}
\caption{\textsc{Sparse sampling problem in PAT in circular geometry.} The induced  acoustic pressure
is  measured at $M$ detector locations on the boundary of the  disc $B_R$ indicated  by white dots in the left image.  Every detector  at location $z_m$ measures  a time dependent pressure signal $\pp[m, \edot]$, corresponding to  a column in the right image.}\label{fig:circular}
\end{centering}
\end{figure}
\end{psfrags}

\subsection{Discretization and sparse sampling}

In practical applications, the acoustic pressure $\wave \source$ can only be
measured with a finite number of acoustic detectors.
The standard sampling scheme  for PAT in circular geometry
 assumes  uniformly sampled  values
\begin{align} \label{eq:data}
     &\pp[m, \edot]
     \coloneqq \wave \source\kl{ z_m, \edot}\,,
    \quad \text{ for }  m = 1, \dots,  M \,,
     \\ \label{eq:sampling}
     & \text{with} \quad z_m
	 \coloneqq
     \begin{bmatrix} R \cos \kl{2\pi(m-1)/M} \\ R\sin \kl{2\pi(m-1)/M}  \end{bmatrix}
      \,.
\end{align}
Here  $\pp[m, \edot] \colon [0,T] \to \R$ is the signal corresponding to the $m$th  detector, and $M$ is the total number of detector locations.
Of course, in practice also the signals $\pp[m, \edot] $ have to be represented by discrete  samples. However, temporal samples can easily be collected at a high sampling rate compared to the spatial sampling, where each sample requires a separate sensor.

In the case that a sufficiently  large number of detectors is used, according to Shannon's sampling theory,
implementations of full data methods yield almost artifact free reconstructions (for a detailed analysis of sampling in PAT see \cite{haltmeier2016sampling}). As the fabrication of an array of detectors is demanding,
experiments using integrating line detectors are often carried out using a single line detector, scanned on circular paths using scanning stages~\cite{NusEtAl10, GruEtAl10}, which is very time consuming.  Recently, systems using arrays of $64$ parallel line detectors have been demonstrated~\cite{gratt201564line, beuermarschallinger2015photacoustic}. To keep production costs low and to allow fast imaging the number $M$ will typically be kept much smaller
than advised by Shannon's sampling theory and one deals with highly under-sampled data.

Due to the high frequency information contained  in time,  there is still hope to recover
high resolution images form spatially under-sampled data. For example, iterative algorithms,
using  TV minimization yield good reconstruction results from undersampled data  (see \cite{scherzer2009variational,meng2012vivo,guo2010compressed,arridge2016accelerated}). However, such algorithms are quite
time consuming  as they require evaluating the   forward and adjoint problem repeatedly
(for TV typically at least several hundreds of times).  Moreover, the reconstruction quality depends  on
certain a-priori assumptions on the
class of objects to be reconstructed such as sparsity of the gradient.
Image reconstruction with a trained CNN is direct  and requires a  smaller  numerical effort
compared to iterative methods.
Further, it does not require an explicit model for the prior knowledge about the objects to be recovered.
Instead,  such a model is implicitly learned  in a data-driven manner based on the training data
by adjusting the weights of the CNN to the provided training data during the training phase.

\section{Deep learning for PAT image reconstruction}
\label{sec:deep}

Suppose that sparsely sampled data of the form \eqref{eq:data}, \eqref{eq:sampling}
are at our disposal. As illustrated in Figure~\ref{fig:approach} in our deep learning approach
we first apply a linear reconstruction procedure to the sparsely sampled data
$(\pp[m, \edot])_{m=1}^{M}$ which  outputs  a discrete image
$\Xin  \in \R^{\Nx \times \Nx}$. According to Shannon's sampling theory
an aliasing free  reconstruction requires $M \geq \pi  \Nx$ detector  positions
 \cite{haltmeier2016sampling}.
However, in practical applications we will have  $M \ll \Nx$, in which case
severe undersampling artifacts appear in the reconstructed image.
To reduce these artifacts, we apply a CNN to the intermediate
reconstruction  which outputs an almost artifact free reconstruction $\Xout  \in \R^{\Nx \times \Nx}$.
How to implement such an approach  is described in the following.

\subsection{Image reconstruction by neural networks}
\label{sec:nn}

The task of  high resolution image reconstruction can be formulated as supervised machine
learning problem. In that context, the aim is finding a restoration function
$\Phi\colon \R^{\Nx \times \Nx}\rightarrow\R^{\Nx \times \Nx}$ that maps
the  input image $\Xin  \in \R^{\Nx \times \Nx}$ (containing undersampling artifacts) to
the output image $\Xout  \in \R^{\Nx \times \Nx}$ which should be almost
artifact free. For constructing such a function $\Phi$, one assumes that  a family of training
data   $\TT  \coloneqq (\Xin_n,\Xout_n)_{n=1}^N$ are given. Any training example
$(\Xin_n,\Xout_n)$ consist of an  input image $\Xin_n$ and
a corresponding artifact-free output image $\Xout_n$.
The restoration function is constructed in such a way that the training error
\begin{equation} \label{eq:err}
 	E(\TT; \Phi) \coloneqq
 	\sum_{n=1}^N d(\Phi(\Xin_n),\Xout_n)
\end{equation}
is minimized, where $d \colon \R^{\Nx \times \Nx} \times \R^{\Nx \times \Nx} \to \R$
measures the error made by the function $\Phi$ on the training examples.

Particular powerful supervised machine learning methods are based on neural networks (NNs).
In such a situation, the restoration function is taken  in the form
\begin{equation} \label{eq:nn}
	 \Phi_{\WW} = (\sigma_L  \circ W_L)  \circ \cdots \circ (\sigma_1  \circ W_1)  \,,
\end{equation}
where  any factor $\sigma_\ell  \circ W_\ell$
is the composition of a linear transformation (or matrix) $ W_\ell\in \R^{D_{\ell+1}\times D_\ell}$ and a
nonlinearity $\sigma_\ell  \colon \R \to \R$  that is applied component-wise.
Here $L$ denotes the number of processing layers, $\sigma_\ell$
are so called activation functions and $\WW := (W_1,\ldots,W_L)$ is the weight vector.
Neural networks can be interpreted to  consist  of several layers,
where the factor $\sigma_\ell  \circ W_\ell$ maps the variables in layer $\ell$
to the variables in layer $\ell+1$. The variables in the first layer
are the entries of the input vector $\Xin$ and the variables in the last layer
are the entries of the output vector $\Xout$. Note that in our situation we have
 an equal number of variables $ D_1  = D_{L+1} = \Nx^2$ in the input and the output layer.
Approximation properties of NNs have been analyzed, for example,
in~\cite{funahashi1989approximative,hornik1991approximation}.

The entries of the weight vector $\WW$ are called weights and  are the variable parameters
in the NN.  They are adjusted during the training phase prior to the actual image reconstruction
process. This is  commonly implemented using
gradient descent methods to minimize the training set error
\cite{bishop2006pattern,goodfellow2016deep}
\begin{equation} \label{eq:err-nn}
	E(\TT,\WW)
	\coloneqq
	E(\TT,\Phi_{\WW})
	= \sum_{n=1}^N d\kl{\Phi_{\WW}(\Xin_n),\Xout_n}
\end{equation}
The standard gradient method uses the update  rule $\WW^{(k+1)}  =  \WW^{(k)}  - \eta  \,  \nabla E(\TT,\WW^{(k)})$, where
 $\nabla E$ denotes the gradient of the error function in the second component and $\WW^{(k)}$ the weight vector in the $k$th iteration.
 In the context of neural networks the update term is also known as error backpropagation.
If the number of training examples is large, then the gradient method becomes slow.
In such a situation, a popular acceleration is the stochastic gradient descent algorithm \cite{bishop2006pattern,goodfellow2016deep}.
Here for each iteration a small subset $\TT^{(k)}$ of the whole training set  is chosen randomly at any iteration and the weights are adjusted
using the modified update formula $\WW^{(k+1)} = \WW^{(k)} -  \eta \,   \nabla E(\TT^{(k)},\WW^{(k)})$ for the $k$th iteration.
In the context of image reconstruction similar acceleration strategies are known as ART or Kaczmarz type reconstruction methods   \cite{decezaro2008steepest,GorBenHer70,natterer2001mathematics}.
The number of elements in $\TT^{(k)}$ is called batch size and  $\eta$ is referred to as the learning rate. To stabilize the iterative  process, it is common to add a so-called momentum term  $\beta \, (\WW^{(k)} -\WW^{(k-1)})$ with some nonnegative parameter $\beta$ in the update of  the $k$th iteration.

\subsection{CNNs and the U-net}
\label{sec:cnn}

In our application, the inputs and outputs are high dimensional vectors.
Such large-scale problems  require special network  designs, where the weight matrices
are not taken as  arbitrary  matrices  but  take a special form  reducing its effective dimensionality.
When the input is an image, convolutional neural networks (CNNs)   use such special network designs
that are  widely and successfully used in various applications \cite{lecun1989backpropagation,bishop2006pattern}.
A main property of CNNs is the invariance with respect to  certain transformations of
the input.  In CNNs, the weight matrices are block diagonal, where each block corresponds to  a convolution  with a filter of small support and the number of blocks corresponds to the number of different filters (or channels) used in each layer. Each block is therefore a sparse band matrix,  where the non-zero entries of the band matrices determine the filters of the convolution. CNNs are currently extensively used in image processing
 and image classification, since they outperform most comparable algorithms \cite{goodfellow2016deep}.
They are also the method of choice for the present paper.

\begin{figure}[htb!]
\centering
   \includegraphics[width=\textwidth]{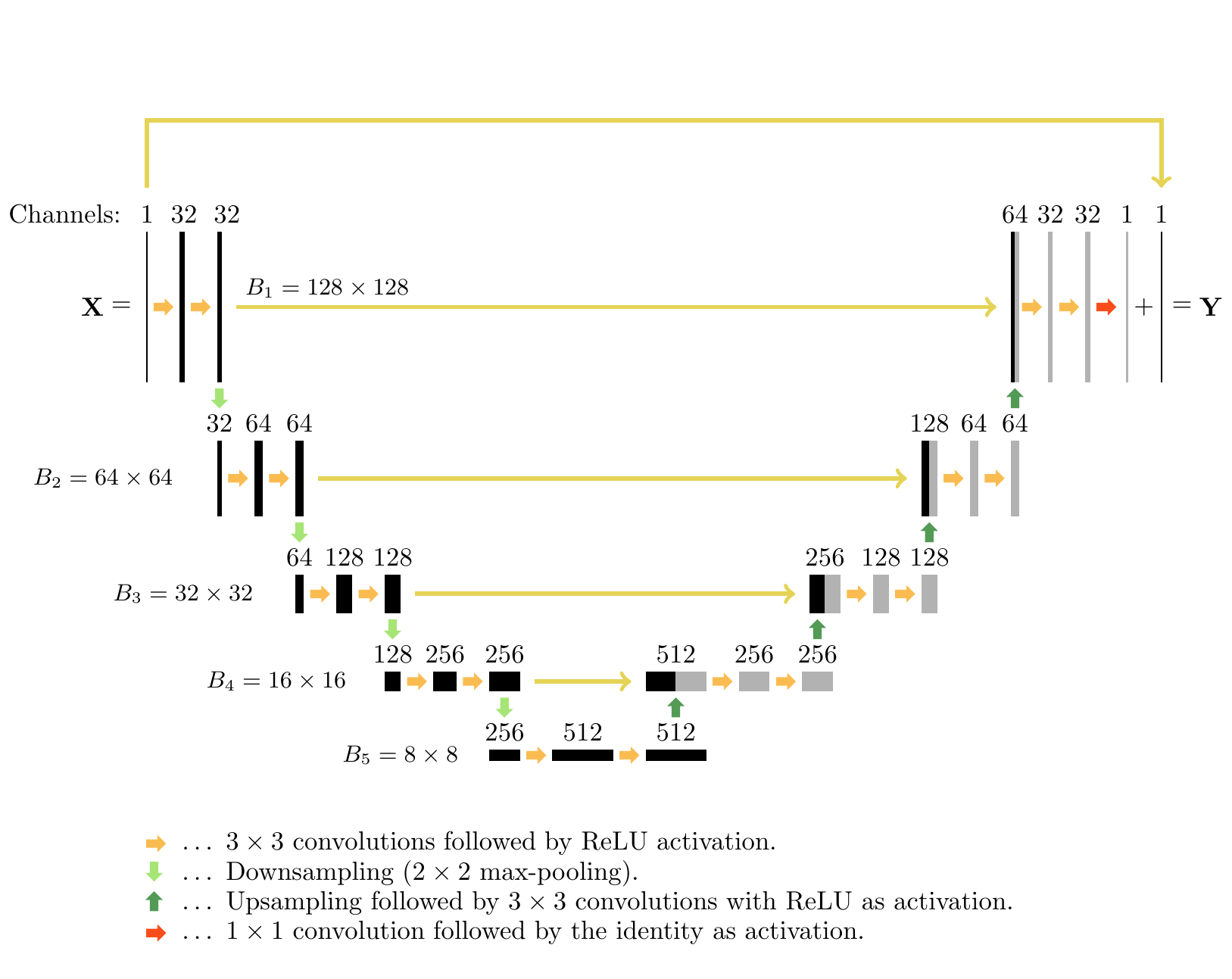}
\caption{\textsc{Architecture of the used CNN.} The\label{fig:net} number written above each layer denotes the number of convolution kernels (channels), which is equal to number of images in each layer. The numbers $B_1, \ldots, B_5$ denote the dimension of the images (the block sizes in the weight matrices), which stays constant in every row. The long yellow arrows indicate direct connections with subsequent concatenation or summation for the upmost arrow.}
\end{figure}

There are various CNN designs that  can differ in the number of layers, the form of the
activation
functions and the particular form of the weight matrices $W_\ell$.
In this paper, we use a particular  CNN based on the so-called U-net introduced in
\cite{ronneberger2015unet}. It has been originally  designed for biomedical image segmentation and recently
been used for low dose CT in~\cite{jin2016deep,han2016deep}. The U-net is based on the so-called fully convolutional network  used in reference  \cite{long2015fully}.
Such network architectures employ multichannel filtering  which means that the weight matrix
in every layer consists of a family of multiple convolution filters followed by the rectified linear unit (ReLU) as activation function. The rectified linear unit is defined by $\operatorname{ReLU}(x)
\coloneqq \max \set{x,0}$.
As shown in~\cite{han2016deep},  the residual images   $\Xin - \Xout$  often have a simpler structure and are
more accessible to the U-net than the original outputs.
Therefore, learning the residuals and subtracting them from the inputs after the last layer is more
effective than directly training  for $\Xout$. Such an approach is followed in our implementation.
The resulting deep neural network architecture is shown in Figure~\ref{fig:net}.

\subsection{PAT using FBP combined with the U-net}
\label{sec:deeppat}

We are now ready to present the proposed deep learning approach for PAT
image reconstruction from sparse data, that uses the FBP algorithm as linear preprocessing step
followed by the U-net for removing undersampling artifacts.
Recall that we have given sparsely sampled data $(\pp[m, \edot])_{m=1}^{M}$
of the form \eqref{eq:data}, \eqref{eq:sampling}. A discrete high resolution
approximation  $\Xout  \in \R^{\Nx \times \Nx}$  with $\Nx \gg M$ of the original
object  is then reconstructed as follows.
\begin{enumerate}[label=(S\arabic*)]
\item\label{alg-S1} Apply the FBP algorithm to   $\pp$ which yields an reconstruction
$\Xin \in \R^{\Nx \times \Nx}$ containing undersampling artifacts.

\item\label{alg-S2} Apply the U-net shown in Figure~\ref{fig:net} to $\Xin$ which yields an image $\Xout \in \R^{\Nx \times \Nx}$
with significantly reduced undersampling artifacts.
\end{enumerate}

The above two steps can also be combined to  a single network with the FBP in the first layer and the U-net for the   remaining layers. 
Note that the first step could also be replaced by another linear
reconstruction  methods such as time reversal and the second step
by a different CNN. Such alternative implementations will be investigated in
future studies.
In this work, we use the FBP algorithm described in \cite{FinHalRak07}  for solving  step \ref{alg-S1}. 
It is based on discretizing  the inversion formula
\eqref{eq:fbp2d}  by replacing the inner  and the outer  
integration by numerical quadrature and  uses an  interpolation procedure to reduce the numerical complexity. For details on the implementation we refer to \cite{FinHalRak07,haltmeier2011mollification}.

A crucial ingredient in the above  deep learning method is the adjustment of the actual weights in the U-net, which
have to be trained on an appropriate training data set.  For that purpose we construct training data $\TT = (\Xin_n, \Xout_n)_{n=1}^N$  by first creating certain phantoms $\Xout_n$.
We then simulate sparse data by  numerically  implementing the well-known solution formula
for the wave equation and subsequently construct $\Xin_n$  by applying the FBP algorithm
of \cite{FinHalRak07} to the sparse data.
For training the network we apply  the stochastic gradient algorithm for minimizing the training set error \eqref{eq:err},
 where we take the error  measure $d$  corresponding to the  $\ell^1$-norm $ \snorm{\Xout}_1  = \sum_{i_1,i_2 =1}^{\Nx} \sabs{ \Xout[i_1,i_2]}$.

\section{Numerical realization}
\label{sec:num}

In this section, we give details on the numerical implementation
of the deep learning approach and present  reconstruction results under various
scenarios.

\subsection{Data generation and network training}

For all numerical results presented below we use $\Nx=128$ for the image size
and take $R=1$ for the radius of the measurement curve. For the
sparse data in \eqref{eq:data} we use $M=30$ detector locations and discretize
the pressure signals $\pp[m, \edot]$ with $300$ uniform samples in the
time interval $[0, 2]$.
In our initial studies, we generate simple phantoms consisting of indicator functions of ellipses
with support in the unit cube $ [-1,1]^2\subseteq \R^2$.
For that purpose, we randomly generate solid ellipses $E$ by sampling independent
uniformly distributed random variables. The centers are selected uniformly  in $(-0.5,0.5)$
and the minor and major axes uniformly in $(0.1,0.2)$.

For the training of the network on the ellipse phantoms we  generate two different data sets, each consisting
of $N = 1000$ training pairs $(\Xin_n, \Xout_n)_{n=1}^{N}$.
One set of training data corresponds
to  pressure data without noise and for the second data set  we added random
noise to the simulated pressure data. The outputs $\Xout_n$ consist of  the
sum of  indicator  functions   of ellipses  generated  randomly as
described above that are sampled on the $128 \times 128$ imaging grid.
The number of ellipses in each training example
is also taken randomly according to  the uniform distribution  on $\set{1, \dots, 5}$.
The input images  are generated numerically by first computing  the sparse
pressure  data using the  solution formula for the wave equation and then applying the FBP algorithm to obtain $\Xin_n$.

For actual training, we  use the stochastic gradient descent algorithm
with a batch size of one for minimizing \eqref{eq:err-nn}.
 We train for  60 epochs which means we make 60 sweeps  through the whole training dataset.
We take  $\eta = 10^{-3}$ for the learning rate, include a momentum parameter
$\beta = 0.99$, and use the mean absolute error  for the distance measure in \eqref{eq:err-nn}.
The weights in the $j$th layer are
initialized by sampling the uniform distribution on $[-H_\ell,H_\ell]$
where $H_\ell \coloneqq \sqrt{6} / \sqrt{D_\ell+D_{\ell+1}}$ and $D_\ell$ is the size of the input in layer $\ell$. This initializer is due to Glorot~\cite{glorot2010}.
We use $\Nf =32$ channels for the first convolution and  the total number of layers
is $\Nl = 19$.

\begin{figure}[thb!]
\centering
	\includegraphics[width=0.225\textwidth, height=0.225\textwidth]{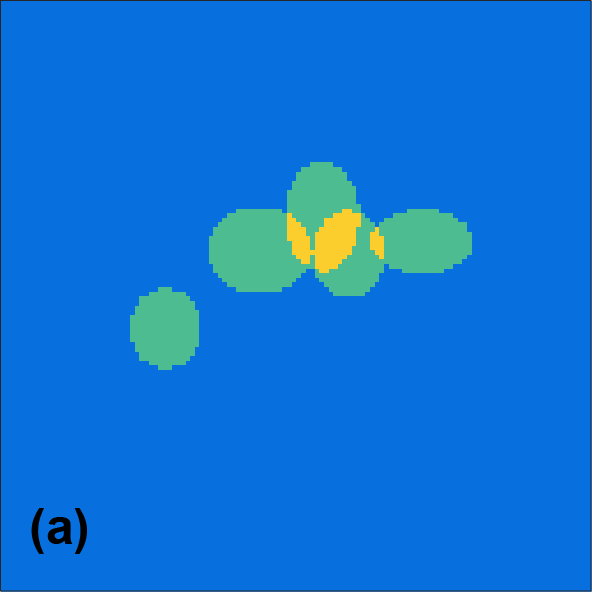}
	\includegraphics[width=0.225\textwidth, height=0.225\textwidth]{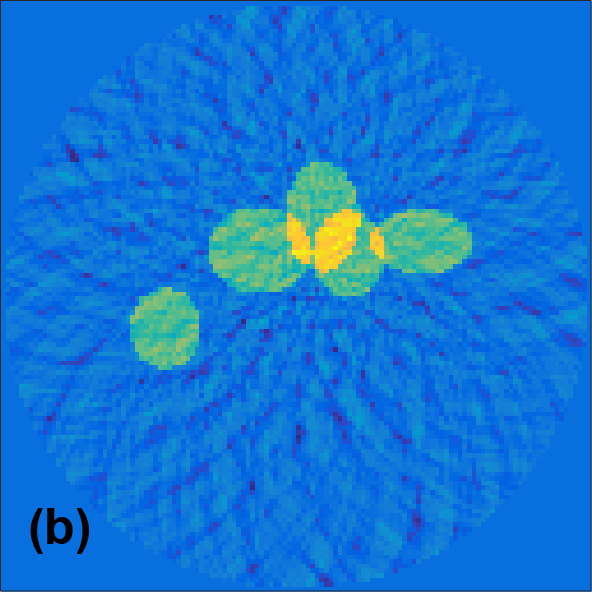}
	\includegraphics[width=0.225\textwidth, height=0.225\textwidth]{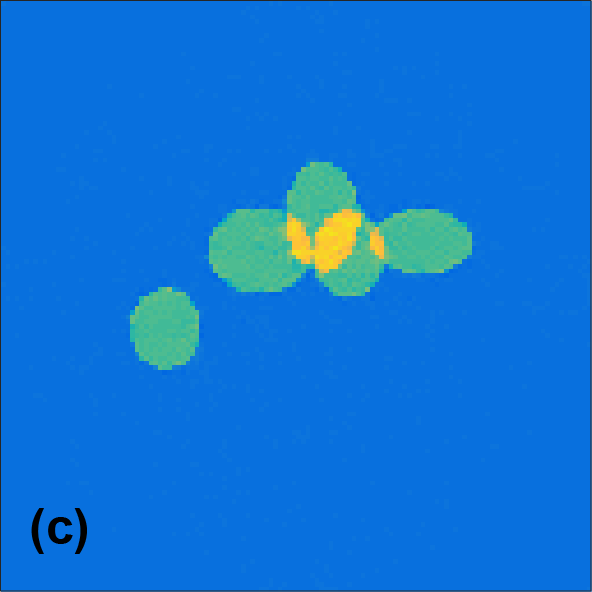}
	\includegraphics[width=0.225\textwidth, height=0.225\textwidth]{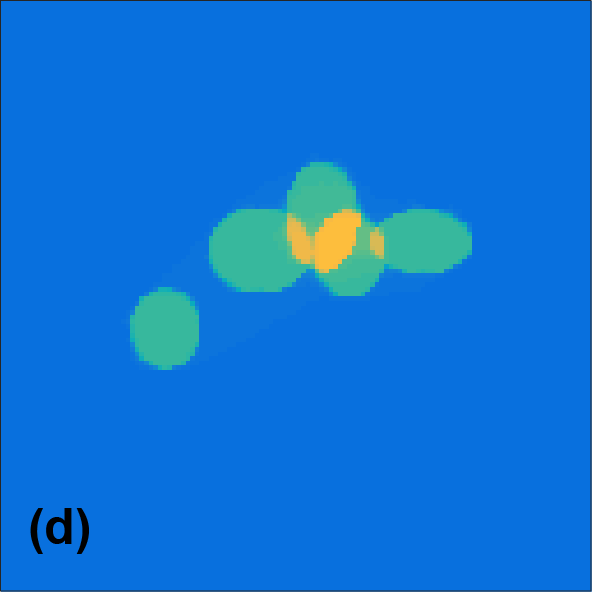}
	\includegraphics[height=0.225\textwidth]{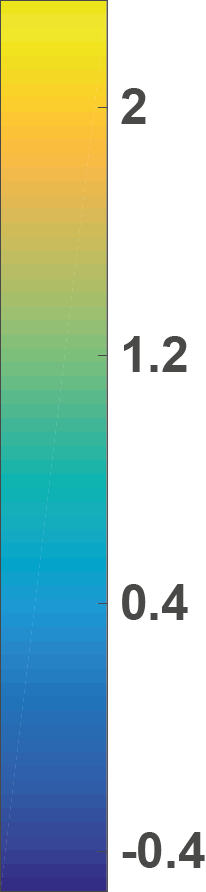}
    \caption{\textsc{Results for simulated data (all images are displayed using the same colormap).}
    (a) Superposition of 5 ellipses as test phantom;
    (b)  FBP reconstruction;
    (c) Reconstruction using the proposed CNN;
    (d) TV reconstruction.}
    \label{fig:exact}
\end{figure}

\subsection{Numerical results}

We first test the network trained above on a test set of 50 pairs $(\Xin, \Xout)$
that  are generated according to the  random model for the training data described above.
For such random ellipse phantoms, the  trained network is in all tested
case able to almost completely eliminate the sparse data artifacts in the test images.
Figure~\ref{fig:exact} illustrates such results for one of the test phantoms.
Figure~\ref{fig:exact}(a) shows the phantom,    Figure~\ref{fig:exact}(b) the  result of the FPB algorithm
which contains severe undersampling artifacts and Figure~\ref{fig:exact}(c)  the result of applying the CNN (right) which is
almost artifact  free. The actual relative $\ell^2$-reconstruction error
$\snorm{\Xout_{\rm CNN}  - \Xout}_2/\snorm{\Xout}_2$  of the CNN reconstruction
is $0.0087$ which is much smaller than the relative error of FBP
 reconstruction  which is 0.1811.

We also compared our trained network to penalized TV minimization
\cite{scherzer2009variational,acar1994analysis}
\begin{equation}\label{eq:tv}
\frac{1}{2}
\snorm{ \pp - \waved  (\Xout)}_2^2 + \lambda \snorm{\Xout }_{\rm TV} \to \min_{\Xout } \,.
\end{equation}
Here  $\waved$ is a discretization of the PAT forward operator using $M$ detector locations and $\Nx$ spatial discretization points and  $\snorm{\edot}_{\rm TV}$ is the discrete total variation.
For the  presented results, we take  $\lambda = 0.002$ and used the lagged diffusivity algorithm
\cite{vogel1996} with 20 outer and 20 inner iterations for numerically minimizing
\eqref{eq:tv}.  TV minimization exploits the sparsity of the gradient as prior information and therefore is especially
well suited for reconstructing sums of indicator functions and can be seen as state of the art  approach for
reconstructing such type  of objects. As can be seen from the results~ in Figure~\ref{fig:exact}(d),
 TV minimization  in fact gives very  accurate results.  Nevertheless, the deep learning approach yields
 comparable results in  both cases. In terms of the relative $\ell^2$-reconstruction error,  the CNN
 reconstruction even outperforms the  TV reconstruction (compare with Table~\ref{tab:comparison}).

\begin{figure}[thb!]
\centering
	\includegraphics[width=0.225\textwidth, height=0.225\textwidth]{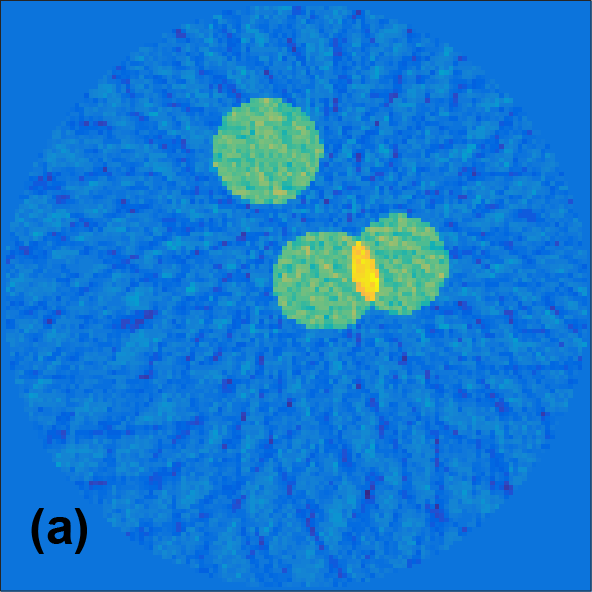}
	\includegraphics[width=0.225\textwidth, height=0.225\textwidth]{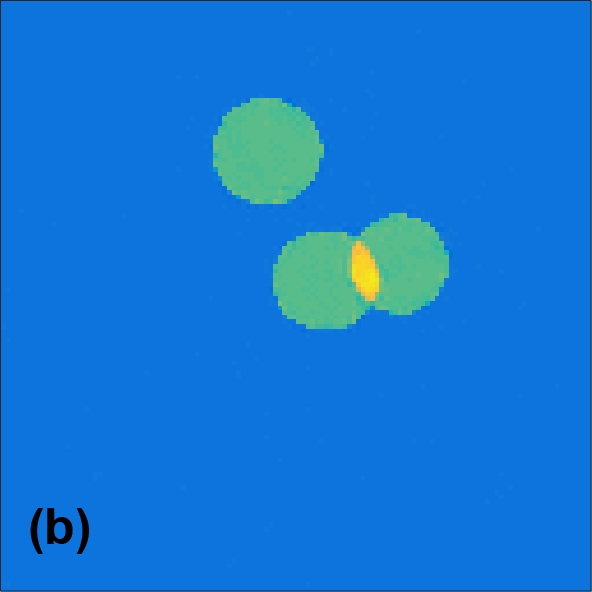}
	\includegraphics[width=0.225\textwidth, height=0.225\textwidth]{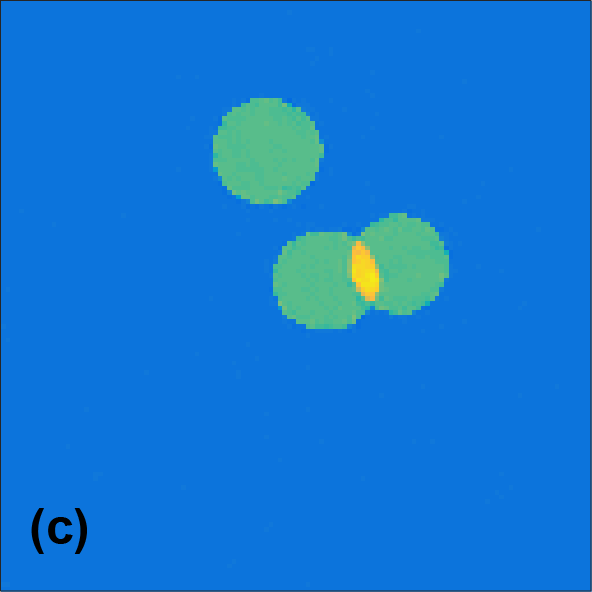}
	\includegraphics[width=0.225\textwidth, height=0.225\textwidth]{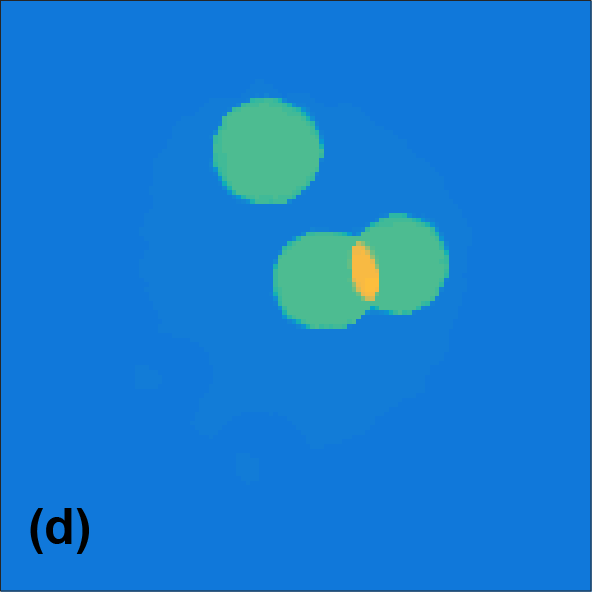}
	\includegraphics[height=0.225\textwidth]{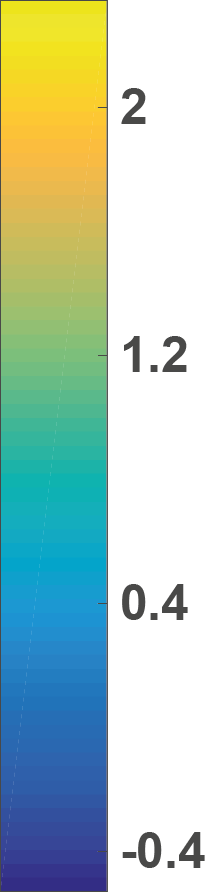}
    \caption{\textsc{Results for noisy test data with  2\% Gaussian noise added (all images are displayed using the same colormap).}
    (a) Reconstruction using the FBP algorithm;
    (b) Reconstruction using the CNN trained without noise;
    (c) Reconstruction using the  CNN trained on noisy images;
    (d) TV reconstruction.}\label{fig:noisy}
\end{figure}

In order to test the stability  with respect to noise we also test the network on reconstructions coming from noisy data.
For that purpose, we added Gaussian noise  with a standard deviation equal to 2\% of the maximal value to
simulated pressure data. Reconstruction results are shown  in Figure~\ref{fig:noisy}. There we show  reconstruction
results with two differently trained networks. For the results shown in Figure~\ref{fig:noisy}(b)
the CNN has been trained on the exact data, and for the results shown in Figure~\ref{fig:noisy}(c)
it has been trained on noisy data. The reconstructions using  each of the networks are again
almost artifact free. The reconstruction from the same data with  TV minimization
is shown in Figure~\ref{fig:noisy}(d).  The relative $\ell^2$-reconstruction errors for all reconstructions
are given in Table~\ref{tab:comparison}.

\begin{figure}[thb!]
\centering
    \includegraphics[width=\textwidth]{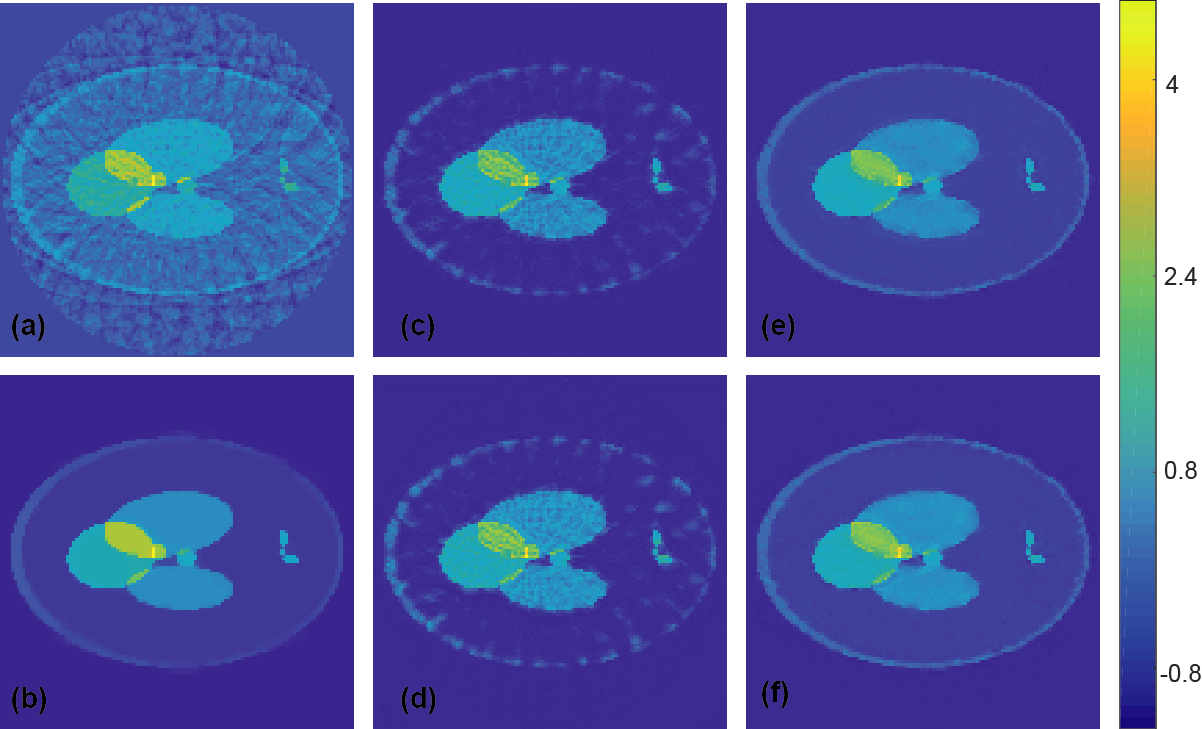}
    \caption{\textsc{Reconstruction results   for a Shepp-Logan type phantom from data with  2\% Gaussian noise added  
    (all images are displayed using the same colormap).}
    (a) FBP reconstruction;
    (b) Reconstruction using TV minimization.
    (c) Proposed CNN using wrong training data without noise added;
    (d) Proposed CNN using wrong training data with noise added;
    (e) Proposed CNN using appropriate training data without noise added;
    (f) Proposed CNN using appropriate training data with noise added
    }
    \label{fig:shepp}
\end{figure}

\subsection{Results for Shepp-Logan type phantom}

\begin{figure}[thb!]
\centering
    \includegraphics[width=\textwidth]{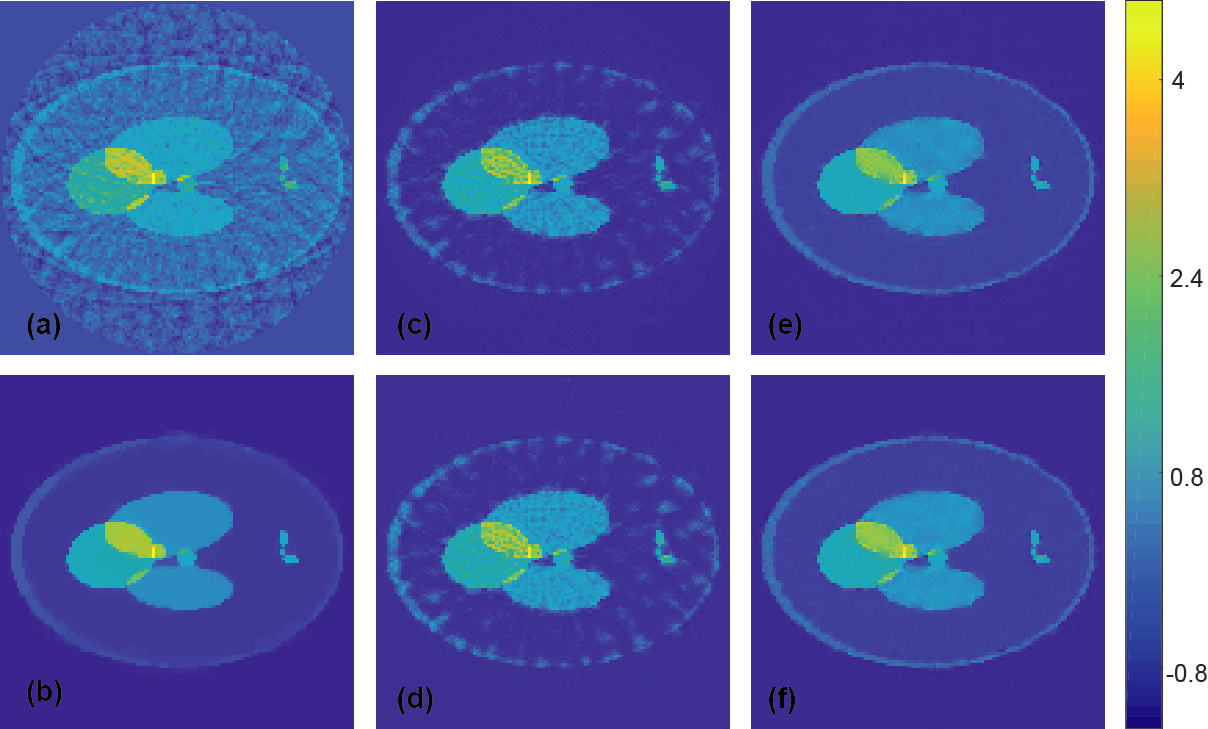}
    \caption{
    \textsc{Reconstruction results   for a Shepp-Logan type phantom using 
    simulated data  
    (all images are displayed using the same colormap).}
    (a) FBP reconstruction;
    (b) Reconstruction using TV minimization.
    (c) Proposed CNN using wrong training data without noise added;
    (d) Proposed CNN using wrong training data with noise added;
    (e) Proposed CNN using appropriate training data without noise added;
    (f) Proposed CNN using appropriate training data with noise added.}
    \label{fig:2shepp}
\end{figure}

In order to investigate the limitations of the proposed  deep learning approach, we additionally  applied the
CNN (trained on the ellipse phantom class) to test phantoms where the  training data are not appropriate (Shepp-Logan type phantoms). 
Reconstruction results for such a  Shepp-Logan type  phantom 
from exact data are shown  in Figure~\ref{fig:shepp}, which compares  results using FBP (Figure~\ref{fig:shepp}(a)), 
TV minimization (Figure~\ref{fig:shepp}(b)) 
and  CNN  improved versions using the ellipse phantom class without noise (Figure~\ref{fig:shepp}(c)) and with 
 noise (Figure~\ref{fig:shepp}(d)) as training data.  
 Figure~\ref{fig:2shepp} shows similar results for noisy measurement data with added Gaussian noise  with a standard deviation equal to 2\% of the maximal 
 pressure value.
 As expected, this time the network does not completely
remove all artifacts. However, despite  the Shepp-Logan type test object has features not appearing in the training data,  still many artifacts are removed by the network trained on the ellipse phantom class.

We point out, that the less good performance of CNN in Figure~\ref{fig:shepp}(a)-(d) and~\ref{fig:2shepp}(a)-(d)  is due to the non-appropriate training data and not due to the type of phantoms or the CNN approach itself. To support this claim,
we trained  additional CNNs on the union of 1000 randomly generated ellipse phantoms and 1000 randomly generated Shepp-Logan type phantoms. The Shepp-Logan type phantoms have position, angle, shape and intensity of every ellipse chosen uniformly at random  under the side constraints that the
 support of every ellipse  lies inside the unit disc.
The results  of the CNN trained on the new training data  and are shown in Figures~\ref{fig:shepp} (e), (f) for exact measurement data and in Figure~\ref{fig:2shepp} 
(e), (f) for noisy measurement data. For both results we applied a CNN trained using training data without (e) and with noise (f).  And indeed, when using appropriate training data including Shepp-Logan typ phantoms, the CNN is again comparable to TV minimization.
We see these results quite  encouraging;  future work will be done to extensively test 
the framework using a variety of training and test data sets, including real world data.

The  relative $\ell^2$-reconstruction errors  for all presented numerical results are summarized in Table~\ref{tab:comparison}.

\begin{table}[thb!]\centering
\begin{tabular}{l | l l  l  l l l}
\toprule
phantom      &  FBP &   TV &  ELL   &   ELLn  &   SL   &   SLn 
\\
\midrule
5 ellipses  (exact)  &     0.1811 & 0.0144  &    0.0087   &  - & - & -
\\
3 ellipses (noisy)  & 0.1952   &  0.0110      &   0.0051    & 0.0038 & - & - 
\\
Shepp-Logan (exact) &   0.3986   & 0.0139    &     0.1017        &  0.1013  & 0.0168 & 0.0186
\\
Shepp-Logan (noisy) &   0.3889   & 0.0154    &     0.1054       &  0.1027  & 0.0198 & 0.0206
\\
\bottomrule
\end{tabular}\medskip
\caption{\textbf{Relative\label{tab:comparison} $\ell^2$-reconstruction  errors  for the four different test cases}. 
Compared are FBP, TV,  
and  the proposed CNN reconstruction trained on  the  class of ellipse phantoms without noise (ELL) and with noise (ELLn), 
 as well as trained on  a class containing Shepp-Logan type phantom without noise (SL) and with noise (SLn).}
\end{table}

\subsection{Discussion}

The above results demonstrate that deep learning  based  methods are a promising tool to improve PAT image reconstruction. The presented results show that appropriately trained CNNs can 
significantly reduce under sampling artifacts and increase reconstruction quality. To further support this claim, in Table~\ref{tab:average} we show the averaged relative $\ell^2$  reconstruction error  for 100 Shepp-Logan type phantoms (similar to the ones in Figure~\ref{fig:shepp}. We see that even in the case where we train the network for the different  class of ellipse shape phantoms, the error decreases significantly compared to FBP.

\begin{table}[thb!]\centering
\begin{tabular}{l | l l  l  l l l}
\toprule
phantom      &  FBP &   TV &  ELL   &   ELLn  &   SL   &   SLn 
\\
\midrule
Exact   &    0.2188   &    0.0140    &     0.1218        &   0.1078  & 0.0215& 0.0213
\\
Noisy &   0.2374 &  0.0150    &      0.1290       &  0.1141  & 0.0256 & 0.0243
\\
\bottomrule
\end{tabular}\medskip
\caption{\textbf{Average error analysis}. \label{tab:average} 
Relative $\ell^2$-reconstruction error averaged of over 
 100 Shepp-Logan type phantoms. Compared are FBP, TV, 
 proposed CNN reconstruction trained on  a class of ellipse phantoms without noise (ELL) and with noise (ELLn), 
 as well as trained on  a class of containing Shepp-Logan type phantom without SL  and with noise (SLn).}
\end{table}

Any reconstruction  method for solving an ill-posed underdetermined problem, either implicitly or explicitly  requires 
a-priori knowledge about the objects to be reconstructed. In classical variational regularization, a-priori knowledge is incorporated by  selecting an element which has minimal (or at least small value)  of a regularization functional among all objects consistent with the data. In the case of  TV, this means that phantoms  with small total variation 
($\ell^1$ norm of gradient) are  reconstructed.  
On the other hand,  in deep learning based reconstruction methods a-priori knowledge is not explicitly prescribed in advance.  
Instead, the a-priori knowledge in encoded in the given training 
class and CNNs are trained to   automatically learn the structure  of desirable outputs.     In the above results, the  training class consists of piecewise constant  phantoms  which have small total variation. Consequently, TV regularization  is expected to perform well. It is therefore not surprising, that in this case TV minimization outperforms CNN based methods.  However, the CNN based methods are more  flexible  in the sense that  by changing the training set they can be adjusted to very different type of phantoms. For example, one can train the  CNN for classes of  experimental PAT data, where it may be difficult to find an appropriate convex regularizer.

\subsection{Computational efforts}

Application of the trained CNN for image reconstruction is non-iterative and efficient.
In fact, one application of the used CNN requires  $\mathcal{O}(\Nf^2  \Nl \Nx^2)$ FLOPS, where $\Nf$ is the number of channels for the first convolution and $\Nl$ describes the depth of the network. Moreover, CNNs are easily accessible to parallelization. For high resolution images, $\Nf^2 \Nl$ will be in the order of  $\Nx$ and therefore the effort for
one evaluation of the CNN is comparable to effort of one evaluation of the PAT forward operator and its adjoint,
which both require  $\mathcal{O}(\Nx^3)$ FLOPS. However, for computing the minimizer of \eqref{eq:tv} we have to repeatedly evaluate the PAT forward operator and its adjoint.  In the examples presented above, for TV minimization
we evaluated both operations 400 times,  and therefore  the deep learning image reconstruction approach is expected  to be  faster than TV minimization or related iterative image reconstruction approaches.

For training and evaluation of the U-net  we use the Keras software (see \url{https://keras.io/}), which is a high-level application programming interface written in Python. Keras runs on top of TensorFlow (see \url{https://www.tensorflow.org/}), the open-source software library for machine intelligence. These software packages allow an efficient and simple implementation of the modified U-net according to Figure~\ref{fig:net}. The filtered backprojection and the TV-minimization have been implemented in MATLAB. We perform our computations using an Intel Core i7-6850K CPU and a Nvidia Geforce 1080 Ti GPU.
The training time for the CNN using the training set of 1000 ellipse phantoms has been 16 seconds per epoch,
 yields 16 minutes for the overall training time (using 60 epochs). For the larger mixed training data set
(consisting of 1000 ellipse phantoms and 1000  Shepp-Logan type phantoms) one epoch requires 25 seconds.
Recovering a single image requires  15 milliseconds for the  FBP algorithm and 5 milliseconds  for applying the CNN.
The reconstruction time for the TV-minimization (with 20 outer and 20 inner iterations) algorithm has been 25 seconds.
In summary, the  total reconstruction time using the two-stage deep learning approach is 20  milliseconds, which is over 1000 times faster than the time required for the TV minimization algorithm. Of course, the reconstruction times  strongly depend on the implementation of
TV-minimization algorithm and the implementation of the CNN approach.  However, any  step in the iterative TV-minimization has to be evaluated in a sequential manner, which is a conceptual limitation of iterative methods. Evaluation of the CNN, on the other hand, is non-iterative and
inherently parallel, which allows efficient parallel GPU computations.

\section{Conclusion}
\label{sec:discussion}

In this paper, we developed a deep learning approach for PAT from sparse data.
In our approach, we first  apply a linear reconstruction algorithm to the
sparsely sampled data and subsequently apply a CNN with weights adjusted   to
a set of training data. Evaluation of the  CNN is non-iterative and has a similar numerical effort
as the standard FBP algorithm for PAT.   The proposed deep learning  image reconstruction approach
has been shown to offer a reconstruction quality similar to state of the art iterative algorithms
and at the same time requires a computational effort  similar to direct algorithms such as FBP.
The presented numerical results can be seen as a proof of principle, that deep learning
is feasible and highly promising  for image reconstruction  in PAT.

As demonstrated in Section~\ref{sec:num} the proposed deep learning framework already
offers a reconstruction quality comparable to  state of the art iterative  algorithms for the sparse
data problem in PAT. However, as  illustrated by Figures~\ref{fig:shepp}  and Figures~\ref{fig:2shepp}  this requires the
PAT image to share similarities with the training data used to adjust the weights of the CNN.
In future work, we will therefore investigate and test our approach under various real-world scenarios
including realistic phantom classes for training and testing, different measurement geometries,  and  increased
discretization sizes. In particular, we will also train and evaluate the CNNs  on real world data.

Note that the results in the present paper assume an  ideal impulse response of the acoustic measurement system. For example, this is appropriate for  PAT using   integrating optical line detectors, which have broad detection bandwidth; see \cite{paltauf2007photacoustic}. In the case that  piezoelectric sensors are used for acoustic detection, the  limited bandwidth 
is an important issue that  must  be taken into account in the PAT forward model and the PAT inverse problem \cite{paltauf2017piezoelectric}. 
In particular, in this case, the CNN must also be trained to learn a deconvolution process addressing the  impulse response  function. Such investigations, as well as the application to real  data, are beyond the scope of this  paper and  have been addressed in our very recent work \cite{schwab2018dalnet}. (Note that the presented paper has already been submitted much earlier, initially in April 14, 2017 and to IPSE in July 1, 2017.)

It is an interesting line of future research using other CNNs that may outperform  the currently implemented  U-net.  We further work on the theoretical analysis of
our proposal providing insight why it works that well, and how to steer the network design for further improving its performance and flexibility.

\end{document}